\documentclass[a4paper,conference]{IEEEtran}
\usepackage{graphicx}
\usepackage{comment}
\usepackage{amsmath,amssymb} 
\usepackage{color}

\usepackage{epsfig}
\usepackage{graphicx}
\usepackage{amsmath}
\usepackage{amssymb}
\usepackage{booktabs}
\usepackage{multirow}
\usepackage{mathrsfs}
\usepackage{bm}
\usepackage{cite}
\usepackage[font=small,skip=5pt]{caption}
\usepackage{floatrow}

\ifCLASSINFOpdf
\else
\fi

\begin{document}
%
\title{Augmented Bi-path Network for Few-shot Learning}



%
\author{\IEEEauthorblockN{
Baoming Yan\IEEEauthorrefmark{1}\IEEEauthorrefmark{2}\IEEEauthorrefmark{3},
Chen Zhou\IEEEauthorrefmark{1}\IEEEauthorrefmark{2},
Bo Zhao\IEEEauthorrefmark{4},
Kan Guo\IEEEauthorrefmark{3},
Jiang Yang\IEEEauthorrefmark{3},
Xiaobo Li\IEEEauthorrefmark{3},
Ming Zhang\IEEEauthorrefmark{2} and
Yizhou Wang\IEEEauthorrefmark{2}}
\IEEEauthorblockA{\IEEEauthorrefmark{2} Peking University, \IEEEauthorrefmark{3} Alibaba Group, \IEEEauthorrefmark{4} The University of Edinburgh}
\IEEEauthorblockA{\footnotesize{\{bmyan, zhouch18, yizhou.wang\}@pku.edu.cn; mzhang@net.pku.edu.cn; \{guokan.gk, yangjiang.yj, xiaobo.lixb\}@alibaba-inc.com; bo.zhao@ed.ac.uk}}}


\maketitle

\begin{abstract}
Few-shot Learning (FSL) which aims to learn from few labeled training data is becoming a popular research topic, due to the expensive labeling cost in many real-world applications. One kind of successful FSL method learns to compare the testing (query) image and training (support) image by simply concatenating the features of two images and feeding it into the neural network. However, with few labeled data in each class, the neural network has difficulty in learning or comparing the local features of two images. Such simple image-level comparison may cause serious mis-classification. To solve this problem, we propose Augmented Bi-path Network (ABNet) for learning to compare both global and local features on multi-scales. Specifically, the salient patches are extracted and embedded as the local features for every image. Then, the model learns to augment the features for better robustness. Finally, the model learns to compare global and local features separately, \emph{i.e.}, in two paths, before merging the similarities. Extensive experiments show that the proposed ABNet outperforms the state-of-the-art methods. Both quantitative and visual ablation studies are provided to verify that the proposed modules lead to more precise comparison results.
\end{abstract}

\ifCLASSOPTIONpeerreview
\begin{center} \bfseries EDICS Category: 3-BBND \end{center}
\fi
%
\IEEEpeerreviewmaketitle

\section{Introduction}
In recent years, Deep Learning methods have achieve significant progress in computer vision by applying deeper architectures \cite{Krizhevsky2012,simonyan2014very,Szegedy2015,He2016} to bigger datasets \cite{deng2009imagenet,lin2014microsoft,OpenImages2}. When training a deep neural network, the performance heavily depends on the amount of labeled training data. However, in many real-world tasks, it is time-consuming even prohibitive to collect and annotate enough data for training the popular deep networks. For example, annotating some fine-grained categories \cite{WelinderEtal2010} or medical data \cite{gulshan2016development} are restricted by not only the few available samples but also the few domain specialists. Therefore, how to get rid of cumbersome labelling and train a good classification model with few labeled data, \emph{i.e.}, Few-shot Learning (FSL) \cite{fe2003bayesian,Koch2015,Snell2017,Sung2018,Zhao2018MSplit}, is a valuable research problem.

\footnote{\noindent{\IEEEauthorrefmark{1} Equal contribution.}
}

Many methods have been proposed to deal with the Few-shot Learning problem in past decades. Early studies \cite{fe2003bayesian,fei2006one} use a small number of samples to directly construct a model for classifying new samples. However, the model is not able to learn the real data distribution or generalize to testing data. 
Recently, Meta-learning based methods \cite{Vinyals2016,Snell2017,Sung2018} are proposed, in which many episodes (basic tasks) are sampled from the training data. The model is trained on many sampled tasks for learning meta-knowledge that is generalized to a distribution of tasks. Methods in this framework differ in the design of the classifier for basic tasks. \cite{Snell2017} used a prototype (class center) based nearest neighbor classifier to classify testing (query) images based on those training (support) images. \cite{Sung2018} proposed to learn to compare the similarity between the query and support images. With the shared feature embedding network, the features of two images are concatenated and fed it into the comparison neural network which outputs the similarity between two images. 

\begin{figure}[t]
{\caption{\footnotesize{The effectiveness of salient patches. S stands for the training (support) image. Q1 and Q2 are testing (query) images. In previous methods, only the global features (of the whole images) are compared, as shown in (a). However, our method compares both global and local features by extracting the  salient image patches. Hence, the comparison result is more precise.}}\label{Fig1}}
{\includegraphics[width=8.0cm]{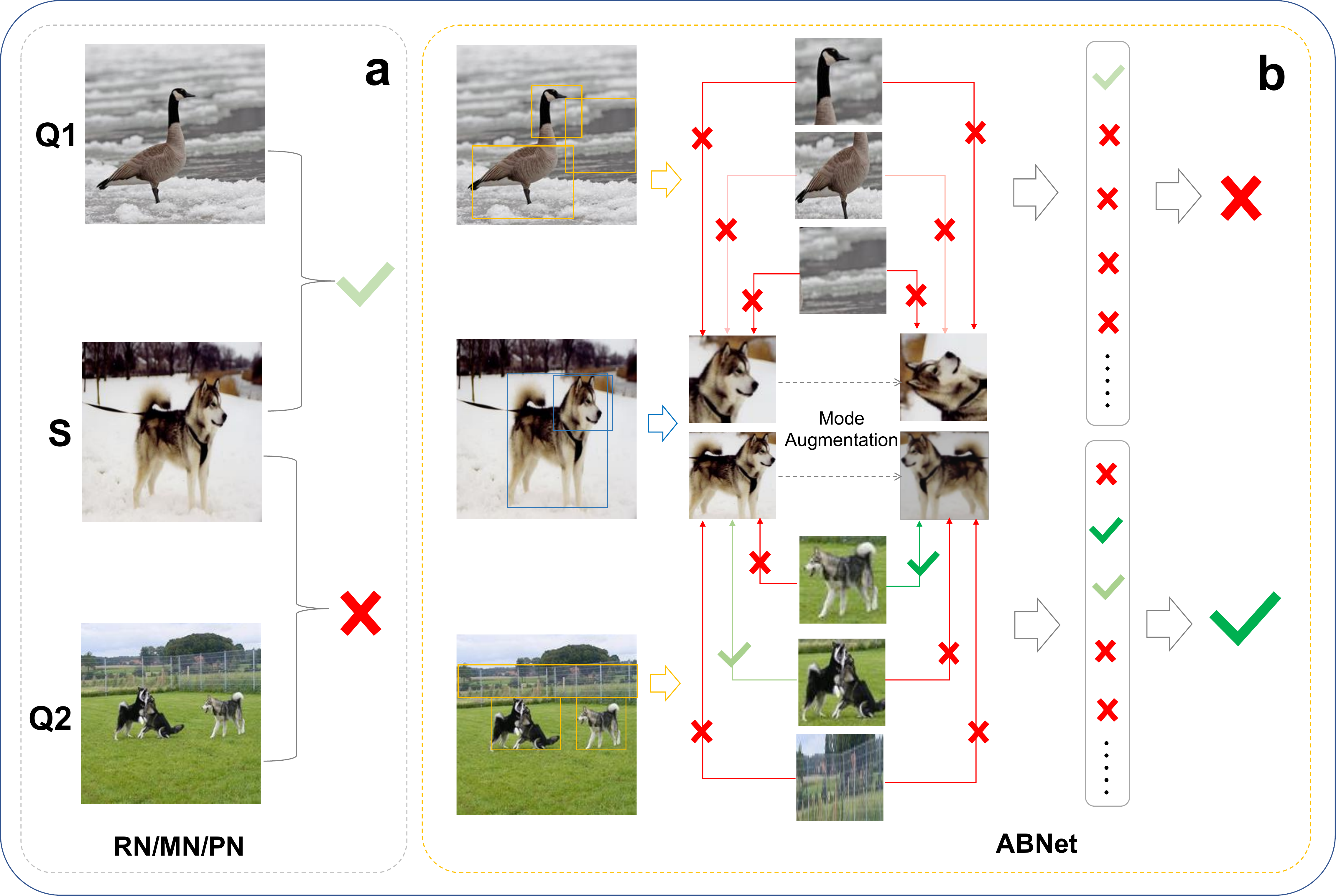}}

\end{figure}

However, with few labeled data in every class, the neural network has difficulty in learning or comparing the local features of two images. Such simple image-level comparison may cause serious mis-classification. As illustrated in Fig.~\ref{Fig1}, given one support image (S) and two query images (Q1 and Q2), the global features of S and Q1 are more similar than those of S and Q2. However, S and Q2 belongs to the class ``dog'', while Q1 is an image of a ``bird''. If we extract and compare some salient image patches, \emph{e.g.}, head and legs, we can easily find that S and Q2 have more similar patches, and they should be classified to the same class.

For more precise comparison, we propose Augmented Bi-path Network (ABNet) to learn to compare both global and local features on multi-scales. Our method includes four main modules. First, the salient patches that contains informative parts are extracted for every image. Second, the original image and its salient patches are embedded by the shared feature embedding module. Third, we learn to augment both global and local (salient patch) features of support images for better robustness. Fourth, we learn to compare the features of support and query images and output the similarity between the two images. Instead of calculating the similarity based on concatenated features directly \cite{Sung2018}, we generate the similarity maps based on concatenated features and learn to re-weight them. Then the global and local similarity maps are merged to produce the overall similarity. Extensive experiments on three challenging benchmarks show that our method outperforms the state-of-the-art methods with a large margin. Ablation studies are provided for verifying that the proposed modules are important for achieving better performance. Visual analysis is given for better illustration of how feature augmentation and local features lead to correct classification.

The main contributions of this paper includes two folds:
\begin{itemize}
\item We propose Augmented Bi-path Network (ABNet) to learn to compare both global and local features of support and query images, which includes two novel modules, \emph{namely}, ``Learning to Augment'' and ``Learning to Compare''. \\
\item We evaluate our approach on three challenging Few-shot Learning benchmarks, miniImageNet, Caltech-256 and tieredImageNet. Our ABNet outperforms the state-of-the-art by a large margin.
\end{itemize}


\section{Related work}
\subsection{Meta-learning Based Methods}
Meta-learning based methods \cite{Koch2015,Vinyals2016,Snell2017,Munkhdalai2017,Santoro2016,Zhou2018} 
aim to learn a more generalized model by meta-training on many sampled FSL tasks. 
Typically, Finn et al.\cite{Finn2017} propose an model-agnostic algorithm for meta-learning that trains a model’s parameters such that a small number of gradient updates will lead to fast learning on a new task In addition, Ravi et al.\cite{Ravi2017} describe an LSTM-based model for meta-learning. The model is trained to discover a good initialization of the learner’s parameters, as well as a successful mechanism for updating the parameters to new task. 

The idea of ``learn to compare'' is also widely used. Gregory Koch et al.\cite{Koch2015} first propose siamese neural networks for One-shot image recognition. Through a shared network structure, deep features are learnt and compared to decide whether the two inputs belong to the same class. Oriol Vinyals et al.\cite{Vinyals2016} build a matching network, which learns a LSTM encoder to embedding the deep features conditioned on the specific support set and query set. Inspired by the evaluation process of Few-shot Learning, they construct similar few-shot classification task in the training data. Following the same training strategy, ProtoNet\cite{Snell2017}, RelationNet \cite{Sung2018} and many other superior networks\cite{Wang2018a,Ravi2017,Nichol2018} are proposed. 

Our ABNet also belongs to meta-learning based methods. The main difference between aforementioned methods and ours is that we learn to compare both global and local similarities of the support and query images instead of only using the global features. We also learn to augment features instead of using hand-craft augmentation strategies.

\subsection{Data Augmentation Based Methods}
One of the important difficulties in Few-shot Learning is the small number of samples. Data augmentation methods 
\cite{Wang2016a,Wang2018a,Schwartz2018,Ren2018,Zhang2018,Chen2018} 
aim to build a generation model to enhance the variety of the input. Eli Schwartz et al.\cite{Schwartz2018} design an auto-encoder\cite{LIOU20083150} for data augmentation. The core idea of their approach is learning to reconstruct a sample via another one, hence the abundant labeled samples can be used to augment the few-shot input. Combining a meta-learner with a hallucinatory, Wang et al.\cite{Wang2018a} present a method to augment the input samples by producing additional imaginary data. Chen et al.\cite{Chen2018} propose to augment the features by semantic information. Creatively in our method, we learn meaningful affine transformation augmentations from training categories in feature map space, and apply them to augment the feature maps of support image.

\subsection{Salient Patches}
Salient patches, which contains rich vision details of the object, is widely used to improve the discrimination of features in various computer vision tasks, such as fine-grained image  classification\cite{zhang2014part,Zhang2016}, clothes retrieval\cite{liu2016deepfashion}, person re-identifica-tion \cite{Zheng2019}, image caption\cite{Gan2016} et al. For these application scenarios, predefined salient patches for specific categories could be extracted by the part detector or key-point detector. While in the few-shot scenario, extracting salient patches for new categories is challenging in the situation of few labeled images and lack of prior knowledge. Wang et al.\cite{Wang2017} employ the semantic embedding of class tag to generate various local features, and combine them into an image-level feature for Few-shot Learning, which could be limited by the performance of unified visual-semantic embedding. Zhang et al. \cite{zhang2019few} employed a saliency network pre-trained on MSRA-B to obtain the foregrounds and backgrounds of images, then hallucinated additional datapoints by foreground-background combinations. Chu et al. \cite{chu2019spot} propose a sampling method based on maximum entropy reinforcement learning to extract various sequences of patches, and aggregate the extracted features for classification. Training the sampler from scratch without effective supervision would extract many background patches and degrades the performance of salient patches. 
In this paper, we learn to compare by exploiting more \emph{class-relevant} salient patches, and learn to augment the samples by learn \emph{intra-class variations} in feature space, which is significantly different from those related works.

\section{Augmented Bi-path Network}
For few-shot classification, the dataset contains a meta-train set and a meta-test set, which have disjoint categories ($C_{meta-train} \cap C_{meta-test}=\varnothing$). The meta-train set, in which each category contains many labeled samples, is used to train a generalized model. Then, it is evaluated on the meta-test set with only few labeled images in each category. In the popular setting, many testing episodes are performed during evaluation process. Each testing episode contains $N$ categories and each category has $K$ labeled images. These labeled data form the support set $S=\{(\mathbf{x}_{1,1}, y_{1,1}), (\mathbf{x}_{1,2}, y_{1,2}), ..., (\mathbf{x}_{N,K}, y_{N,K})\}$, where $(\mathbf{x}_{i,j}, y_{i,j})$ is $j$th datum of $i$th class. The rest unlabeled testing data of these $N$ categories form the query set $Q=\{\mathbf{x}_{1,K+1}, \mathbf{x}_{1,K+2}, ..., \mathbf{x}_{N,K+1}, \mathbf{x}_{N,K+2}, ...\}$. Such setting is called $N$-way $K$-shot learning.

\begin{figure*}[t]
  \centering
  \includegraphics[width=0.95\linewidth]{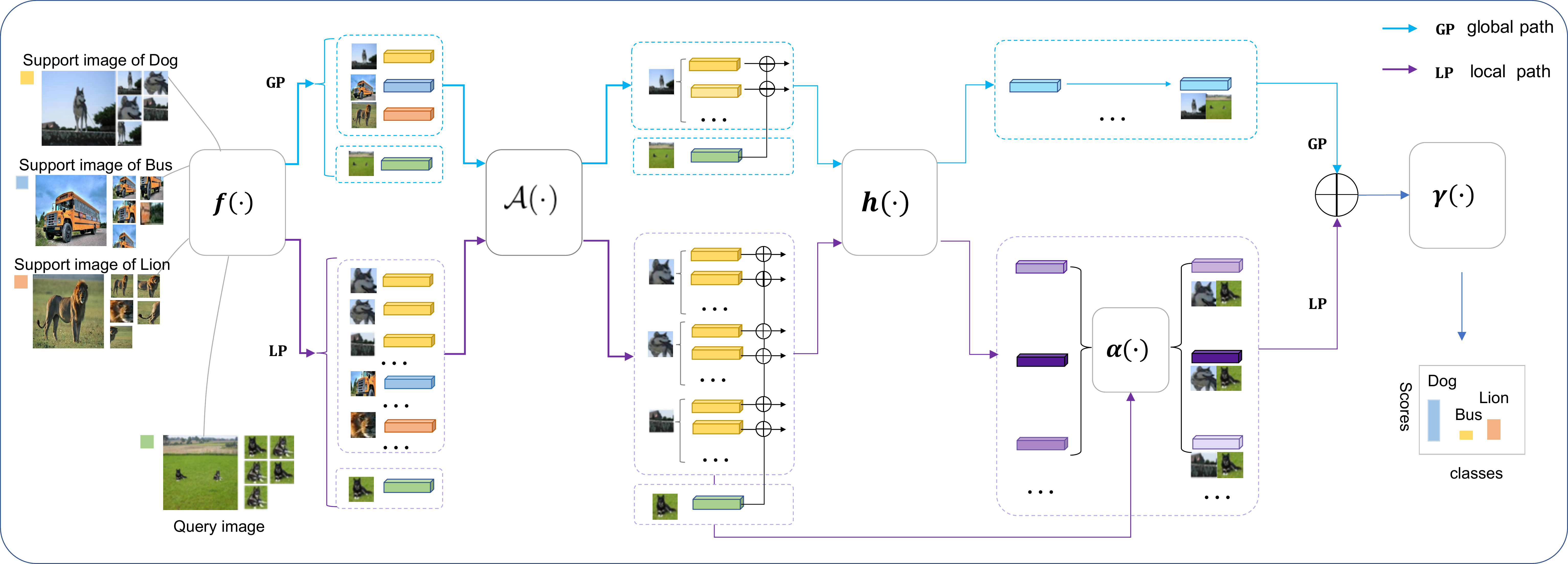}
  \caption{\footnotesize{Illustration of the proposed Augmented Bi-path Network, which contains a global path (GP) for comparing the whole images and a local path (LP) for comparing the extracted salient patches. The whole image accompanied with its salient patches are fed into the network $f(\cdot)$ for feature embedding. To enhance the robustness, a learnable feature augmentation module $\mathcal{A}(\cdot)$ is utilized to augment the support features. Then, similarity maps between augmented support features and the query feature are computed through $h(\cdot)$. Specially for the local path, an re-weighting module $\alpha(\cdot)$ is employed to suppress irrelevant or meaningless patches. Finally, combining GP and LP, the merging module $\gamma(\cdot)$ learns to merge the global and local similarity maps and regress the overall similarity. The label of the query image is predicted as the one with the largest similarity.}}
  \label{Fig2}
\end{figure*}

We follow the widely used episode-based training strategy \cite{Vinyals2016}, which mimics the evaluation process. 
Similar as each testing episode, a training episode is constructed by random sampling $N$ classes from $C_{meta-train}$. $K$ labeled data of every class sever as the support set, and $M$ labeled data from the rest form the query set. The popular methods \cite{Vinyals2016,Snell2017,Sung2018} train a deep embedding model by feeding the whole image into the deep neural network. 
To fully exploit the few labeled images, we propose Augmented Bi-path Network (ABNet) for Few-shot Learning, the framework is illustrated in Fig.~\ref{Fig2}. 
In total, ABNet includes four modules: 
1) Given the support $s$ and query $q$ images, a patch extraction module is first applied to obtain $N$ salient patches for each image. 
2) Then, both the whole image and patches are fed into a shared convolution neural network (CNN) $f(\cdot)$ for feature embedding. 
3) To enhance the robustness, a learnable feature augmentation module is utilized to augment the features of support images by mimicking the diversity of query images. 
4) We learn to compare the features of the support and query images. The similarity maps between two features from the support and query images are computed and re-weighted by a learnable attention module. Finally, combining the global path and local path, the merging module learns to merge the global and local similarity and regress the similarity score of the support and query images. The following subsections are details of the proposed method.

\subsection{Salient Patch Extraction}
Salient patches are vital for the comprehensive description of an object, especially in the few-shot scenario. 
Hence, we develop a salient patch extraction module. Our extraction module starts with the patches sampled by selective search (SS) method \cite{uijlings2013selective}, which applies bottom-up grouping procedure to generate good object locations capturing all scales. Then elaborate selection method is utilized to distill salient patches. 

We measure the importance of patches from two aspects, namely, geometry property and visual salience. Geometry property is referred to the area and aspect ratio of the patch. As small patches usually lack enough discriminate features and large patches are close to the global image, only patches with moderate scale and aspect ratio are significant and will be kept. We use rectangular function $\Pi(x)$ as the geometry property, if the geometry property satisfies the requirement, $\Pi(r_i) = 1$ else $\Pi(r_i) = 0$. 
Visual salience measures the attraction of pixels to human attention, which is the principal element of importance. The Minimum Barrier Distance (MBD)\cite{zhang2015minimum}, which represents the connectivity to the background regions, is utilized to measure the visual salience. Hence, the salience of specific patch could be represented as the average of the pixels in it. 
Taking both geometry property and visual salience into consideration, the total importance of patch $r_i$ is defined as follows\cite{zhang2015minimum}:
\begin{equation}
\begin{aligned}
    S(r_i) &= \Pi(r_i) \cdot \frac{1}{K}\sum_{j=0}^{K}v(p_j)\\
    v(p_j) &= \min_{\pi\in\mathbb{S}}[\max_{t=0}^TI(\pi(t))-\min_{t=0}^TI(\pi(t))]
\end{aligned}
\end{equation}
where $v(p_j)$ is the visual salience of the pixel, $K$ is the total number of pixels in patch $r_i$, $I(\cdot)$ is the pixel value, $\pi$ is a sequence of pixels where consecutive pairs of pixels are adjacent and the total number of pixels is $T$, $\mathbb{S}$ is the set of all sequences that connect $p_j$ and seed pixels from background. 

Then, the patches could be ranked by the importance, and top $N$ patches could be selected to balance efficiency and effectiveness. Our salient patch extraction is not time consuming as the procedure is performed only once and offline before training. As shown in Fig.~\ref{Fig2}, the whole image accompanied with the extracted salient patches introduces multi-scale inputs to the feature embedding module (CNN), which improves its representation ability.

\subsection{Feature Embedding}
Both the whole image and salient patches are fed into the shared CNN backbone $f$ for feature embedding.
Following the classic setting, $f$ consists of four basic convolution blocks. Each block includes a 2D convolution layer with 64 (3x3) kernels, a batch normalization layer and a ReLU nonlinear function. 2x2 max-pooling layer is added after the first two basic convolution blocks. 
Given image $I$, we concatenate the feature maps in the last convolution layer and construct a 3D feature map:
\begin{equation}
    f(I) = \omega_1 \oplus \omega_2 \oplus \cdots \oplus \omega_n,
\end{equation}
where $\omega_i$ is the $i${th} feature map in the output of CNN backbone and $\oplus$ denotes the concatenate operator, $f(I)\in \mathbb{R}^{H\times W\times C}$ with width $W$, height $H$ and $C$ channels. Different from existing methods \cite{Snell2017, bertinetto2018meta}, where feature embedding is reshaped into one dimensional vector as the input of classifiers, we keep the spatial information in the feature map by leveraging the 3D feature map. As training samples are limited in Few-shot Learning, the additional spatial information will benefit the learning with few samples.

\subsection{Learning to Augment}
In Few-shot Learning, the number of training samples is quite limited, while the testing images varies considerably in orientation, viewpoint and clutter background. Hence, even the feature embedding of the same object will be diverse, which results in mis-classification due to the extremely large intra-class distance. To alleviate the problem, we manage to augment the feature embedding of every whole support image $I_s$ and its salient patches in the support set by learning intra-class variations from the categories with sufficient labeled training data. The feature augmentation module aims to learn to transform the support feature to new ones which capture the diversity of query images. Hence, we introduce a regularization between the generated features and the query features. We factorize the complicated variations into several independent modes, each mode is represented by an affine transformation matrix. For a specific point $(x_i, y_i, z_i)$ in a feature map $f(I)$, the transformed location $(x_i^a, y_i^a, z_i^a)$ in the augmented feature map is defined in the equation:
\begin{equation}
\scriptsize
\left(\begin{array}{l}
    x_i^a  \\
    y_i^a \\
    z_i^a
\end{array}\right)= \mathcal{A} \left(\begin{array}{l}
    x_i  \\
    y_i \\
    z_i \\
    1
\end{array}\right)
= \left[\begin{array}{cccc}
    \mathcal{A}_{11},& \mathcal{A}_{12},& \mathcal{A}_{13},&\mathcal{A}_{14}  \\
    \mathcal{A}_{21},& \mathcal{A}_{22},& \mathcal{A}_{23},&\mathcal{A}_{24} \\
    \mathcal{A}_{31},& \mathcal{A}_{32}, &\mathcal{A}_{33},&\mathcal{A}_{34}\\
    0,& 0, &0,& 1
\end{array}\right]\cdot \left(\begin{array}{c}
    x_i  \\
    y_i \\
    z_i \\
    1
\end{array}\right)
\end{equation}
where $\mathcal{A}$ is an affine transformation matrix with learnable parameters. Different from existing methods\cite{Vinyals2016,Santoro2016}, we learn to augment feature instead of using hand-crafted augmentation strategy.  All the feature maps in support set share the same transformation defined by $\mathcal{A}$. A new group of transformed output feature maps $f_k(I_s)$ could be generated, which could be translated, scaled, rotated or affined, whatever.

Hence, employing $K$ affine transformation matrices, the original feature embedding $f(I_s)$ could be greatly expanded into $K+1$ variants defined as $\Omega$: 
 \begin{equation}
 \begin{aligned}
     \Omega = \mathbb{A} f(I_s) = &\{f_{0}(I_s), f_{1}(I_s),\ldots, f_{K}(I_s) \}\\
     \mathbb{A}=&\{\mathcal{A}_0, \mathcal{A}_1,\ldots,  \mathcal{A}_K \}
\end{aligned}
 \end{equation}
where $\mathbb{A}$ is a group of learned affine transformation matrices ($\mathcal{A}_0$ is an identity matrix, indicating the original feature embedding, and $f_{0}(I_s)$ represents $f(I_s)$).

\subsection{Learning to Compare}
Inspired by \cite{Sung2018}, we classify a query image by learning to compare its feature with those of (labeled) support images.
Instead of only comparing the support and query image, we propose to compare both the whole image and salient patches. First, we compare the whole image of the support and query image. Then, we compare the $N$ salient patches of the support and query image. Hence, there are $1+N^2$ comparison results.

\paragraph{Generating Similarity Maps.}
For comparing two features from support and query images ($I_s$ and $I_q$), we first concatenate the support feature and every feature variant ($\Omega_i$) of the support image
in the channel dimension, which is defined as follows:
\begin{equation}
    p^{k}_{q,s} = f(I_q)\oplus f_{k}(I_s)
\end{equation}
Then, a convolution block $g(\cdot)$ is utilized to calculate the similarity and generate the similarity map $g(p^{k}_{q,s})$. Each feature embedding in $\Omega$ is processed independently in this stage. 
All of the $K+1$ similarity maps are concatenated in the channel dimension, and another convolution block $h(\cdot)$ is applied to accumulate the results for calculating the total similarity maps $m_{s,q}$ for $I_s$ and $I_q$:
\begin{equation}
    m_{s,q}=h[g_\phi(p^{0}_{q,s})\oplus g(p^{1}_{q,s})\oplus\cdots\oplus g(p^{K}_{q,s})]
\end{equation}

Similar as the global feature comparison, the comparison between every local patches of support and query image is calculated. In total, $N^2+1$ similarity maps are generated:
\begin{equation}
    \mathbb{G} = \{m_{s,q}\} \cup \{m_{s,q}^{i,j}|i,j \leq N\}
\end{equation}
where $i,j$ is positive integer indicating the index of patches, and $N$ is total number of salient patches. In total, we have one global similarity map and $N^2$ local similarity maps.

\paragraph{Learning to Re-weight} 
Instead of using the original similarity maps directly, we learn to re-weight the local similarity maps for emphasizing class-relevant patches and suppressing background patches. The re-weighting value is adaptively predicted by an attention module $\alpha(\cdot)$, which outputs a normalized weight between 0 and 1.

The input of the attention network is a triplet defined as $(f(I_s),f(I_q), m_{s,q}^{i,j})$, which combines the features of the support and query images and corresponding similarity map. For simply, we use $\alpha(i,j)$ to denote the weight $\alpha(f(I_s),f(I_q), m_{s,q}^{i,j})$.
By multiplying the weight $\alpha(i,j)$ to corresponding local similarity maps $m_{s,q}^{i,j}$, a re-weighted similarity group $\mathbb{G^*}$ could be obtained:
\begin{equation}
        \mathbb{G^*} = \{m_{s,q}\} \cup \{\alpha(i,j) \cdot m_{s,q}^{i,j}|i,j \leq N\}.
\end{equation}

\paragraph{Learning to Merge}
To obtain the overall similarity score between the support and query images, we merge both the global and local similarity maps. All the elements from the weighted similarity group $\mathbb{G^*}$ are concatenated in channel dimension, and regarded as the input to the merge module $\gamma(\cdot)$. The module is utilized to merge these similarity maps and develop a similarity metric, which contains two convolution blocks and two fully-connected layers. The overall similarity $o(s,q)$ between $I_q$ and $I_s$ is defined as:
\begin{equation}
    o(s,q) = \gamma\Big([\overset{i,j\leq N}{\underset{i,j=1}\bigoplus} \alpha(i,j) \cdot m_{s,q}^{i,j}] \oplus m_{s,q}\Big),
\end{equation}
where $\oplus$ denotes the concatenate operator.
For few-shot classification, we compute the overall similarities between a query image and all support images. Then, the similarities are averaged for every class. The label is predicted as the one with the largest averaged similarity. 

\subsection{Loss Function} 
To train all the parameters in our model, we minimize the classification loss between the predicted similarity score and ground-truth score. The predicted score $o(i,j)$ is first converted to probabilistic score by Sigmoid function:
\begin{equation}
    P(i,j) = \frac{1}{1 + e^{-o(i,j)}}
\end{equation}
Then the classification loss function is computed as :
\begin{equation}
\mathcal{L}_{cls} = \frac{1}{B\times C}\sum_{i=1}^B\sum_{j=1}^C\left(P(i,j) - \mathbb I(y_i == y_j)\right)^2,
\end{equation}
where $B$ and $C$ are the numbers of query and support images respectively. $y_i$ and $y_j$ are the class labels of the query ($I_q$) and support ($I_s$) images. $\mathbb I(x)$ is an indicator function, $\mathbb I(x)=1$ if $x$ is true and 0 otherwise.

To learn sparse attention values for salient patches, we add the $L_1$ regularization. The loss is defined as follows:
\begin{equation}
    \mathcal{L}_{att} = \frac{S_{att}}{N^2}\sum_{i=1}^{ N}\sum_{j=1}^{N}|\alpha(i,j)|,
\end{equation}
where $N$ is the number of salient patches, $S_{att}$ is scaling factor to make the loss in the same scale.

Besides, we introduce the augmentation loss to regularize the feature augmentation. The feature augmentation module tries to learn the diversity of query images and generate new features with such diversity. Hence, we introduce a regularization between the generated features and the query features. During training, we have labels of both support and query images. For the query image $I_q$, if it has the same label as the support image $I_s$, we compute the mean square error (MSE) between $f(I_q)$ and augmented features $\{f_{k}(I_s)\}_{1\leq k\leq K}$:
\begin{equation}
    \mathcal{L}_{aug} = \frac{S_{aug}}{B \times C}\sum_{i=1}^B\sum_{j=1}^C\sum_{k=1}^{K}(f(I_q)-f_{k}(I_s))^2\mathbb I(y_i == y_j),
\end{equation}
where $K$ is the total number of augmented features and $S_{aug}$ is the scaling factor for augmentation loss. 

Finally, the total loss is computed as the weighted sum of the above losses:
\begin{equation}
\label{loss_fn}
    \mathcal{L} = \mathcal{L}_{cls} + \lambda_{att} \cdot \mathcal{L}_{att} + \lambda_{aug} \cdot \mathcal{L}_{aug},
\end{equation}
where $\lambda_{att}$ and $\lambda_{aug}$ are the weights of the attention loss and augmentation loss.

\section{Experiment}
In this section, we first compare to the state-of-the-art methods on three popular datasets. Then, we show that our method is generalized and can achieve good results on different learning settings. Latter, blation study with visualization is provided to illustrate that every proposed module is important.

\subsection{Datasets}
We do experiments on three datasets, namely, miniImageNet \cite{Vinyals2016}, Caltech-256\cite{griffin2007caltech} and tieredImageNet\cite{Ren2018}. miniImageNet is the most popular benchmark for few-shot classification, Caltech-256 is a larger dataset with more categories, and tieredImageNet is a more challenging dataset with test classes that are less similar to training ones.

{\bf miniImageNet.} The dataset contains 100 classes with 600 images per class \cite{Vinyals2016}. Objects in images have variable appearances, positions, viewpoints, poses as well as background clutter and occlusion. We follow the popular split \cite{Ravi2017}, where 64 classes are for training, 16 classes are for validation and the rest 20 classes are for testing. All images are resized to 84 $\times$ 84 size.

{\bf Caltech-256.} This dataset \cite{griffin2007caltech} contains 30,607 images from 256 object categories. These categories are diverse, ranging from grasshopper to tuning fork. We follow the split for FSL which is provided by \cite{Zhou2018}.
The training, validation and testing sets include 150, 56 and 50 classes respectively. The same to miniImageNet, all images in Caltech-256 are resized to 84 $\times$ 84 size.

{\bf tieredImageNet.} This dataset \cite{Ren2018} contains 608 classes (779,165 images) grouped into 34 higher-level nodes from the ImageNet human-curated hierarchy. This set of nodes is partitioned into 20, 6, and 8 disjoint sets of training, validation, and testing nodes, and the corresponding classes constitute the respective meta-sets. All images are resized to 84 $\times$ 84 size.
\begin{table*}
  \caption{\footnotesize{Few-shot classification accuracy (\%) on miniImageNet, Caltech-256 and tieredImageNet datasets. The results are averaged over 600 testing episodes, and the 95\% confidence intervals are reported. We compare to methods using the same \textbf{4-layer} feature embedding module, \emph{i.e.}, $(64\times64\times64\times64)$.}}
  \label{table1}
  \centering
\scriptsize
  \begin{tabular}{llcccccc}
    \toprule
    \multicolumn{1}{l}{\multirow{2}{*}{Method}} &\multicolumn{1}{l}{\multirow{2}{*}{}}&\multicolumn{2}{c}{miniImageNet} & \multicolumn{2}{c}{Caltech-256}& \multicolumn{2}{c}{tieredImageNet}\\
    \cmidrule(r){3-4}
    \cmidrule(r){5-6}
    \cmidrule(r){7-8}
     &    &5way1shot & 5way5shot&5way1shot&5way5shot&5way1shot&5way5shot\\
    \midrule
    MatchingNet\cite{Vinyals2016} & NIPS'16&43.56$\pm$0.84&55.31$\pm$0.73&45.59$\pm$0.77&54.61$\pm$0.73 & 54.02 & 70.11\\
    MetaLSTM\cite{Ravi2017} & ICLR'17&43.44$\pm$0.77&60.60$\pm$0.71&-&-&-&- \\
    MAML\cite{Finn2017} & ICML'17&48.70$\pm$0.84&55.31$\pm$0.73&48.09$\pm$0.83&57.45$\pm$0.84&51.67$\pm$1.81& 70.30$\pm$0.08\\
    MetaNet\cite{Munkhdalai2017} & ICML'17&49.21$\pm$0.96&-&-&-&-&-\\
    ProtoNet\cite{Snell2017} & NIPS'17&49.42$\pm$0.87&68.20$\pm$0.70&-&-&54.28$\pm$0.67&71.42$\pm$0.61\\
    RelationNet\cite{Sung2018} & CVPR'18&50.44$\pm$0.82&65.32$\pm$0.77&56.12$\pm$0.94&73.04$\pm$0.72&54.48$\pm$0.93&71.32$\pm$0.78\\
    CTM\cite{li2019finding} &CVPR'19 & 41.62 &58.77 &-&-&-&-\\
    Spot\&Learn\cite{chu2019spot} &CVPR'19&51.03$\pm$0.78&67.96$\pm$0.71&-&-&-&-\\
    MetaOptNet\cite{lee2019meta} &CVPR'19  &52.87$\pm$0.57&68.76$\pm$0.48 &-&-&54.71$\pm$0.67&71.79$\pm$0.59\\
    \midrule
    ABNet &&{\bf58.12$\pm$0.94}&{\bf72.02$\pm$0.75}&{\bf63.20$\pm$0.99}&{\bf78.42$\pm$0.69} &{\bf62.10$\pm$0.96}&{\bf75.11$\pm$0.78}\\
    \bottomrule
  \end{tabular}
\end{table*}

\subsection{Implementation Details}
During salient patches extraction, the size requirement is set to be 1\%$\sim$50\% and the aspect ratio requirement is set to be 1/3$\sim$3. To further remove duplicate patches, Non-Maximum Suppression (NMS) technique is applied to keep smaller ones. Then, the extracted patches are ranked by the visual salience, and
top five salient patches are selected for each image. For those images with less than five extracted salient patches, we pad the number to five by duplicating. 

For fair comparison, we employ the most widely used 4-layer convolution module \cite{Snell2017,Sung2018,Ravi2017} with 64 filters in each convolution layer as the backbone. The architecture is $(64\times64\times64\times64)$. This embedding module generates 64 (19$\times$19) feature maps for each input image or patch. Unless specified, all experiments are implemented with this 4-layer backbone. We also provide the results with ResNet backbone on miniImageNet. 

In feature augmentation module, four affine transformation matrices are utilized to learn augmentations from the training data. 
The module $g$ and $h$ for generating similarity maps contains the same basic convolution block as the backbone, and an additional 2$\times$2 max-pooling layer is utilized in $h$. 
The triplet input of attention module $\alpha$ is global-average pooled in spatial dimension and further concatenated into an one dimensional vector. Then three fully-connected layers are employed to learn the attention value. Finally, in the merging module $\gamma$, one basic convolution block followed by two fully-connected layers are applied to regress the similarity score. Besides, ReLU is used as the default activation function in all fully-connected layers except the output layer of attention module and merging module, where Sigmoid is used to normalize the score to be $(0, 1)$.
When training shot $>1$, e.g. 5-shot learning, we average the similarity scores of all training shots.

We train the model from scratch, and no extra data or pretrained models are used.  Adam\cite{Kingma2015} optimizer is used during training with initial learning rate 0.01 for feature augmentation module and 0.001 for the others. The learning rate is decayed in half every 100,000 episodes. $\lambda_{att}$ and $\lambda_{aug}$ are both set to be 0.1. Following \cite{Sung2018}, we report the averaged testing accuracy over 600 episodes, and each episode contains $15$ query images of every class.

\begin{table}
    \caption{\footnotesize{Few-shot classification accuracy (\%) on miniImageNet with ResNet backbones.} }
  \label{tab:resnet}
  \centering
  \scriptsize
  \begin{tabular}{llccc}
  \toprule
    \multicolumn{1}{l}{\multirow{2}{*}{Method}} &\multicolumn{1}{l}{\multirow{2}{*}{}}&\multicolumn{1}{c}{\multirow{2}{*}{Backbone}}&\multicolumn{2}{c}{miniImageNet} \\     \cmidrule(r){4-5}
     &    & & 5way1shot & 5way5shot\\ \midrule
RelationNet\cite{Sung2018} &CVPR'18 & ResNet-18& 58.21&74.29 \\
    CTM\cite{li2019finding} &CVPR'19 & ResNet-18&62.05$\pm$0.55&78.63$\pm$0.06\\
    MetaOptNet\cite{lee2019meta} &CVPR'19 & {ResNet-12}&{62.64$\pm$0.61} &{78.63$\pm$0.46}\\
 \midrule
    ABNet & &ResNet-18&\textbf{63.15$\pm$0.63} & \textbf{78.85$\pm$0.56} \\
    \bottomrule
  \end{tabular}
\end{table}

\begin{figure}[htp]
{\caption{\footnotesize{The 5-way $n$-shot classification accuracy on miniImageNet (left) and Caltech-256 (right). The 4-layer backbone is used. Orange points are results of our ABNet, while cyan points are results of RelationNet.}}\label{Fig3}}
{\includegraphics[width=8.0cm]{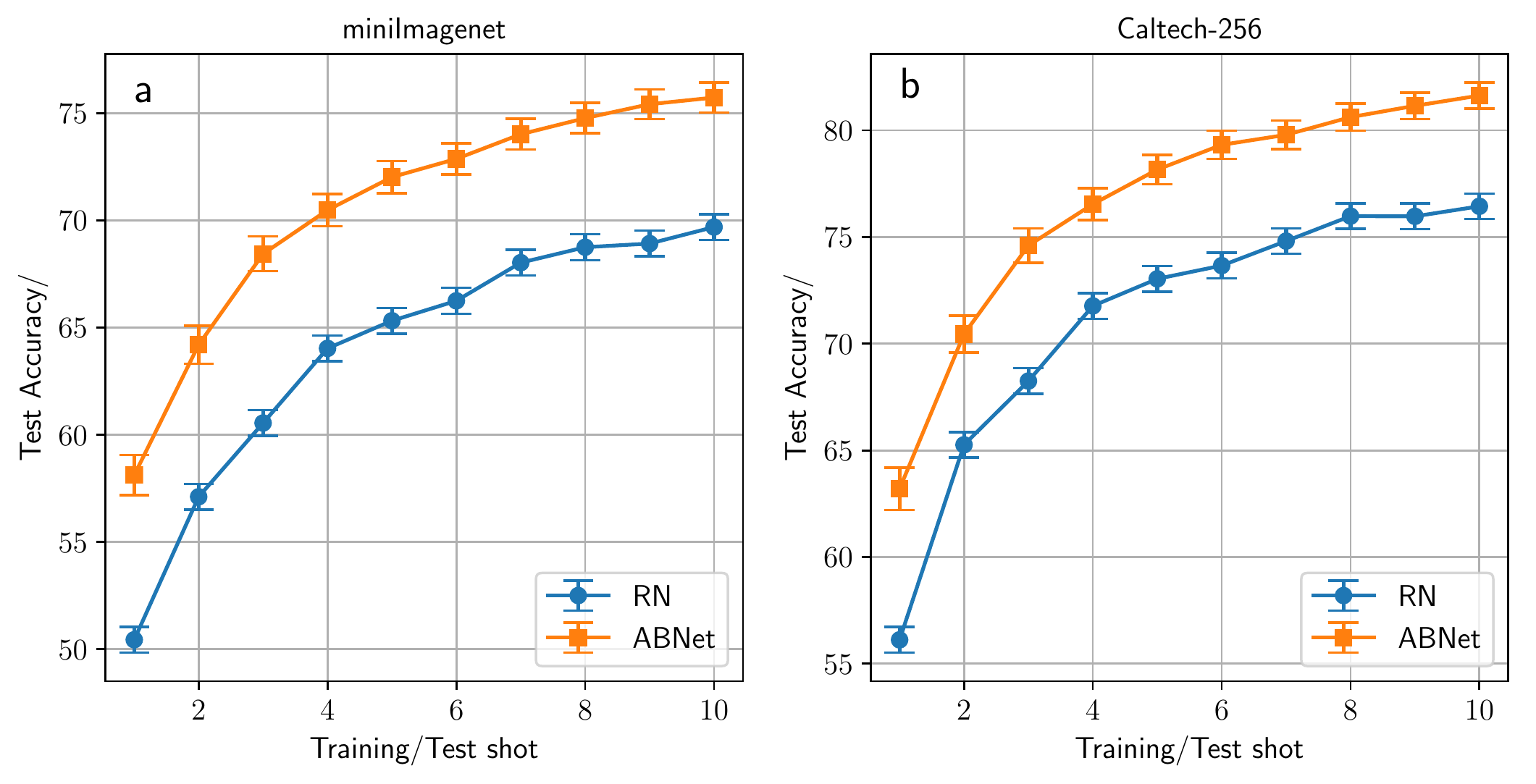}}
\end{figure}

\begin{figure*}[htp]
  \centering
  \includegraphics[width=0.8\linewidth]{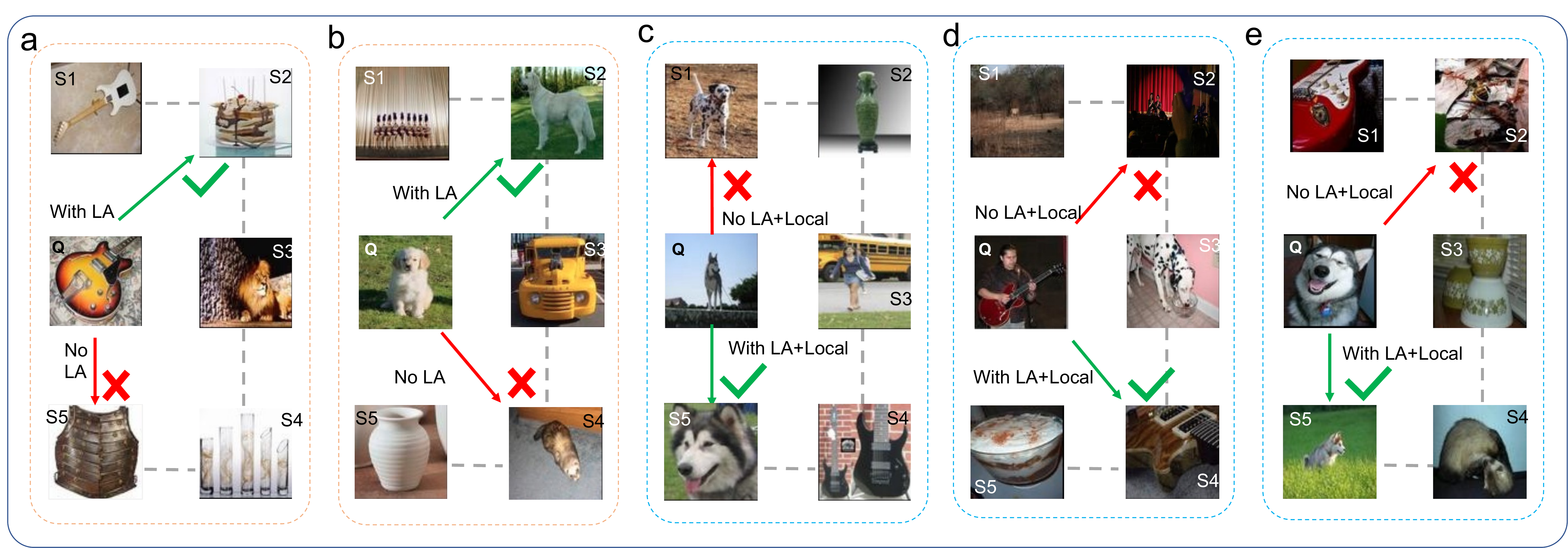}
  \caption{\footnotesize{Visualization of 5-way 1-shot classification results on miniImageNet.
  LA denotes Learning to Augmentation module. Local stands for local features which is produced by  Salient Patch module. Correct predictions are made by the model with feature augmentation when the orientation (a) and viewpoint (b) varies. When the scale of object varies (c), even both scale and viewpoint/orientation varies in (d)/(e), the combination of feature augmentation and local features in ABNet leads to the correct classification.}}
  \label{Fig4}
\end{figure*}
 
\subsection{Comparison to the State-of-the-art}
We compare our approach to several state-of-the-art methods under 5-way-1-shot and 5-way-5-shot settings. The results are shown in Table \ref{table1}. As a deeper backbone with higher resolution input image will always increase the classification performance by a large margin\cite{Mishra2017a,Tan2019}, we compare to methods using the same backbone $(64\times64\times64\times64)$ for fair comparison.

According to Table \ref{table1}, our ABNet achieves the best performance in all settings. 
Our method outperforms the runner-up (MetaOptNet\cite{lee2019meta}) by $5.25\%$ (1-shot) and $3.26\%$ (5-shot) on miniImageNet and $7.39\%$(1-shot) and $3.32\%$(5-shot) on tieredImageNet. In addition, our method also shows significant improvements on Caltech-256 dataset. Compared to RelationNet\cite{Sung2018}, our method achieves $7.08\%$ and $5.38\%$ improvements on 1-shot and 5-shot learning settings respectively. 

Some works employ deeper networks, e.g. ResNet, to extract features. We also provide the results on miniImageNet with ResNet-18 as the backbone in Table \ref{tab:resnet}. Obviously, our method achieves the best performances on two settings. It is interesting that the gaps between different methods (CTM\cite{li2019finding}, MetaOptNet\cite{lee2019meta} and ours) with ResNet backbone are not large. The main reason may be that the features extracted by deep networks are already good enough for few-shot classification.


\subsection{Generalization Ability}
To verify that our method is generalized to settings with different training shots, we provide the detailed comparison to RelationNet\cite{Sung2018} under 5-way-$n$-shot setting where $n\in[1,10]$. As illustrated in Fig.~\ref{Fig3}, along with the increase of training shot, the performance of our ABNet stably improves. On both two datasets, our method always significantly outperforms RelationNet under all settings. It demonstrates that our method has good generalization ability on $n$-shots tasks.



\subsection{Ablation Studies}
To study the effectiveness of Learning to Augment (LA), Salient Patch (SP) and Learning to Re-weight (LR) modules independently, we perform quantitative comparison and visual comparison on miniImageNet dataset.

\noindent {\bf Quantitative comparison.} We start with a baseline model, which only employs global features with handcrafted feature augmentations, e.g.  horizontal flip, rotation of $\pm\pi/2$ and $\pi$. Then we gradually add LA, SP and LR modules. Then, we evaluate the four variants of our method.
\begin{itemize}
    \item Baseline: global features with handcrafted augmentations
    \item Baseline+LA: Baseline with Learning to Augment
    \item Baseline+LA+SP: Baseline with Learning to Augment and Salient Patch
    \item Baseline+LA+SP+LR: Baseline with Learning to Augment, Salient Patch and Learning to Re-weight
\end{itemize}

The classification accuracies for 1-shot and 5-shot learning are evaluated and shown in Table \ref{table2}. Compared to baseline model, the accuracy with Learn to Augment module (Baseline+LA) improves by 1.83\% in 1-shot learning and 1.52\% in 5-shot learning. It means learnable feature augmentations could acquire more appropriate variations from query images.  When Salient Patch model is further introduced (Baseline+LA+SA), the performance is boosted by 1.49\% in 1-shot learning and 1.68\% in 5-shot learning. Hence, the importance of local features is verified. Finally, a significant improvement of 2.36\% and 2.32\% is observed after introducing Learning to Re-weight module (Baseline+LA+SA+LR), which demonstrates the importance of re-weighting local similarity maps before merging. As class-irrelevant patches or background patches could be introduced inevitably by the unsupervised salient patch extraction method, the novel Learning to Re-weight module could emphasize more relevant local similarities by further taking features into consideration.

\begin{table}[htp]
  \caption{\footnotesize{Few-shot classification accuracy (\%) for ablation studies. The results are averaged over 600 test episodes with the 95\% confidence intervals. The baseline model is trained with handcrafted feature augmentation (horizontal flip and rotation), LA: Learning to Augment, SP: Salient Patch, LR: Learning to Re-weight.}}
  \label{table2}
  \centering
  \scriptsize
  \begin{tabular}{ccc}
    \toprule
    \multicolumn{1}{c}{\multirow{2}{*}{Model}} &\multicolumn{2}{c}{miniImageNet}\\
     & 5way1shot & 5way5shot\\
    \midrule
    Baseline &52.44$\pm$0.91&66.50$\pm$0.77\\
    Baseline+LA&54.27$\pm$0.91&68.02$\pm$0.77 \\
    Baseline+LA+SP&55.76$\pm$0.89&69.70$\pm$0.72\\
    Baseline+LA+SP+LR&{\bf58.12$\pm$0.94}&{\bf72.02$\pm$0.75}\\
    \bottomrule
  \end{tabular}
\end{table}

\noindent {\bf Visual Comparison.} The effectiveness of Learning to Augment and Salient Patches is also demonstrated by visualizing the 1-shot classification results in Fig .~\ref{Fig4}. As the orientation and viewpoint of specific object varies in the wild, it is very difficult to predict correct category merely based on one glance (shot) of the object. For example, when the ``guitar'' image in the query set is placed in totally different orientation from the one in support set (see Fig.~\ref{Fig4}(a)), one-shot classifier without feature augmentation fails to recognize it and predicts the wrong category. Similar as the ``dog'' image with different viewpoints in Fig.~\ref{Fig4}(b). However, our method can learn to augment the support image feature and predict the correct category. Moreover, when the scale of object in the query image is significantly different from the one in support image, classifier with merely global features no longer works well and local features are required to make correct prediction. Taking the ``dog'' image as an example, when close-shot image is compared to long-shot image in Fig.~\ref{Fig4}(c), the image of a different specie (category) but similar scale is mis-matched by the method using only global features. In contrast, ABNet with local features could easily recognize the correct specie (category) even the two objects have significantly different scale. Even with different scales, viewpoints (see  Fig.~\ref{Fig4}(d)) and orientations (see Fig.~\ref{Fig4}(e)), ABNet can predict the correct category. The above improvements benefit from the meta-learning ability of Learning to Augment and Learning to Compare modules with salient patches.

\section{Conclusion}
We propose a novel meta-learning based method, namely Augmented Bi-path Network, for Few-shot Learning. The proposed method extends the previous ``learn-to-compare'' based methods by introducing both global and local features on multi-scales. Experimental results show that our method significantly outperforms the state-of-the-art on three challenging datasets under all settings. 
Ablation studies verify the importance of the proposed Learning to Augment, Salient Patch and Learning to Re-weight modules. We also provide visual comparison to illustrate how these modules can improve FSL performance.

\bibliographystyle{IEEEtran}
%
\bibliography{CV_new}



\end{document}